\RequirePackage{amsmath}
\documentclass[natbib]{llncs}
\usepackage{llncsdoc}
\usepackage{graphicx}
\usepackage{caption}
\usepackage[numbers,sort&compress]{natbib}
\usepackage{microtype}
\usepackage{setspace}

\newcommand{\vect}[1]{\mathbf{#1}}
\newcommand{\vects}[1]{\boldsymbol{#1}}
\newcommand{\matr}[1]{\mathbf{#1}}

\newcommand{\jacob}{\mathbf{J}}
\newcommand{\HH}[0]{\mathcal{H}}
\newcommand{\LL}[0]{\mathcal{L}}

\newcommand{\vb}[0]{\vect{b}}

\newcommand{\vh}[0]{\vect{h}}

\newcommand{\vx}[0]{\vect{x}}

\newcommand{\vs}[0]{\vect{s}}

\newcommand{\vsig}[0]{\vects{\sigma}}

\newcommand{\vmu}[0]{\vects{\mu}}
\newcommand{\vepsilon}[0]{\vects{\epsilon}}

\newcommand{\mW}[0]{\matr{W}}

\newcommand{\TT}[0]{\vects{\theta}}
\newcommand{\PP}[0]{\vects{\phi}}
\newcommand{\PS}[0]{\vects{\psi}}

\begin{document}
  \title{Variational Autoencoders for Feature Detection \\ of Magnetic Resonance Imaging Data}
  \titlerunning{VAE for MRI data}
  \author{R. Devon Hjelm \inst{1, 2} \and Sergey M. Plis \inst{2} \and Vince C. Calhoun \inst{1, 2}}
  \institute{University of New Mexico
  \and The MIND Research Network}
  \maketitle

\begin{abstract}
Independent component analysis (ICA), as an approach to the blind source-separation (BSS) problem, has become the de-facto standard in many medical imaging settings. 
Despite successes and a large ongoing research effort, the limitation of ICA to square linear transformations have not been overcome, so that general INFOMAX is still far from being realized.  
As an alternative, we present feature analysis in medical imaging as a problem solved by Helmholtz machines, which include dimensionality reduction and reconstruction of the raw data under the same objective, and which recently have overcome major difficulties in inference and learning with deep and nonlinear configurations.
We demonstrate one approach to training Helmholtz machines, variational auto-encoders (VAE), as a viable approach toward feature extraction with magnetic resonance imaging (MRI) data.
\end{abstract}

\section{Introduction}\label{sec:intro}
Feature selection is a central theme in analyzing many variants of magnetic resonance imaging (MRI) data.  
Supervised approaches that are highly capable of performing regression or classification, but do not rely on features, are at best specialized maps between input data and the output labels. 
They lack the crucial component of ``inference'' to produce generalizations \emph{about} the input data.  
Meanwhile, the inference process and ability to find generalizeable features or structure in the data is at the core of scientific discovery; in MRI research, such structures are necessary for the general goal of understanding the brain.

Inferring the latent structure is generally the goal of unsupervised learning, which has had a wide success in analyzing MRI data. 
When combined with supervised learners, these structures have a diagnostic value.  
Independent component analysis (ICA)~\cite{bell1995} is a representative approach, which has found success as a means for inferring the latent structure in brain imaging data represented via a linear mixture of maximally-independent sources~\cite{smith2009, swanson2010, vince2001}.

While linear mixture models have been very successful in neuroimaging applications, their success relative to nonlinear models \cite{hyvarinen1999nonlinear} is due to simple and tractable inference, not due to a strong belief that linearity is a correct depiction of the latent structure.  
For linear mixtures, non-Gaussian sources are necessary to ensure uniqueness, as for Gaussian sources one cannot guarantee any independence beyond the correlation~\cite{rao1966characterisation}.  
Luckily, the converse is true: non-Gaussian sources with linear mixtures assure maximum independence under a generative learning framework, such as maximum likelihood estimation (MLE)~\cite{dempster1977maximum}.
Requiring the prior distribution be non-Gaussian, while enabling inference and learning with linear ICA methods, breaks down when the relationship between data and sources is nonlinear, necessitating more advanced methods.

Although nonlinear versions of ICA~\cite{ilin2004post, harmeling2003kernel, yang1998information, almeida2003misep}, as well as some alternative nonlinear methods~\cite{hjelm2014restricted} exist, each comes with its shortcomings, and none have been successful enough to supplant linear ICA.  
Alternatively, nonlinear independent component estimation (NICE)~\cite{dinh2014nice} is a method for drawing from a family of nonlinear transformations, $f$, parameterized by feed forward networks (FFNs) such that computing the Jacobian and the inverse are tractable.  
While NICE can estimate sources from nonlinear mixings better than ICA in simulations, it is also limited to square transformations and requires principle component analysis (PCA) to be practical in a medical imaging setting~\cite{castro2016deep}.

In addition to being constrained to square transformations, ICA and many nonlinear variants cannot incorporate multimodal data in a natural way.  
The linear mixing assumption is harder to justify when modes are drawn from fundamentally different distributions, such as MRI, electroencephalography (EEG), and other variables such as age, gender, and clinical diagnoses.

The above issues are ongoing challenges for realizing the full potential of a deep independence network (DIN) or a general INFOMAX approach~\cite{cardoso1997infomax} for feature extraction in medical imaging. 
It is possible that lack of progress has been due to the strong requirement in ICA that the data be the output of a deterministic map of sources.  
As an alternative, we propose learning features in a directed graphical model setting using recent advances in variational inference and demonstrate the effectiveness of this approach with MRI data.

\section{Directed Belief Networks}
Linear mixture models such as ICA fall under a more general category of \emph{volume-preserving bijective maps}~\cite{leondes1998neural}, such that we learn a deterministic parameterized transformation, $f(.; \TT)$, along with a prior distribution of the sources:
\begin{align}
\vh = f(\vx; \TT), \
p_x(\vx) = p_h(f(\vx)) |\jacob|,
\end{align}
where $p_x$ is the density of the data, $p_h$ is the density of the sources, and $\jacob = \det{\partial{f(\vx; \TT)} / \partial{\vx}}$ is the Jacobian. 
For ICA, we have two constraints: first, $\vh = f(\vx) = \mW \vx + \vb$, is a linear transformation with square unmixing matrix, $\mW$, and second, the prior distribution of the sources, $\vh$, is non-Gaussian. 
Probabilistic ICA (PICA) \cite{beckmann2004probabilistic} relaxes the square requirement, but learning is still reliant on a linear transformation as well as a noise operator with known covariance. 
Being a linear transformation, computing the Jacobian, and hence learning, is tractable, but this cannot be said about general nonlinear transformations. 
Nonlinear independent component estimation (NICE) gets around this problem by parameterizing $f$ as a feed-forward network (FFN), such that the affine transformation at each layer is lower or upper triangular, but it is still limited to square transformations.

A directed graphical model or \emph{Bayesian network} is a generative model that represents the density of the data as the marginal of the joint: $p(\vx) = \sum_{\vh} p(\vx, \vh)$, which is composed of a set of prior and conditional distributions that make up an acyclic graph.  
Directed graphical models have been used in various medical imaging settings~\cite{plis2011effective, kim2008hybrid}, but have been limited to relatively simple, often linear configurations. 
A special case of the Bayesian network is the \emph{directed belief network}: a hierarchical model that represents the joint via layers of latent variables that within a layer are conditionally independent:
\begin{align}
p(\vx, \vh) = p(\vx | \vh_1) p(\vh_L) \prod_{l=1}^{L-1} p(\vh_l | \vh_{l+1}),
\end{align}
where $p(\vh_L)$ is the prior distribution of the top or $L$th layer.

Directed graphical models are most commonly trained using the maximum-likelihood estimation (MLE) method, which maximizes the log-likelihood of the data by adjusting parameters of the conditional and prior distributions.\footnote{At least in the parametric case.} 
When present, latent variables need to be marginalized out at each stage during the process; but training can become difficult as marginalizing the joint distribution over the latent variables is often computationally infeasible.  
Potentially, learning can be aided by the use of a posterior, $p(\vh | \vx)$, such that $p(\vx) = p(\vx, \vh) / p(\vh | \vx)$; however, the exact posterior can be equally intractable, particularly when the conditional distributions are complex (e.g., parameterized by highly nonlinear functions). 
%Markov chain monte carlo (MCMC) \cite{neal1992connectionist} can, in principle, be used to find the exact posterior, but it is not practical in large-scale applications due to slow mixing and high computational costs.

\subsection{Variational Inference and Helmholtz Machines}
Some recent advances allow us to more easily train directed graphical models. \emph{Variational inference} makes use of an approximate posterior to compute the variational lower bound of the log-likelihood, $\LL$:
\begin{align}
    \label{eq:approx_logp}
    \log p(\vx) &= \log \sum_{\vh} p(\vx, \vh) 
    = \log \sum_\vh q(\vh|\vx) \frac{p(\vx, \vh)}{q(\vh|\vx)}
    \geq \sum_\vh q(\vh|\vx) \log \frac{p(\vx, \vh)}{q(\vh|\vx)} \nonumber \\
    &\approx \frac{1}{M} \sum_{m=1}^M \log p(\vx, \vh^{(m)}) + \HH(q) := \LL,
\end{align}
where we have used a Monte Carlo estimate for the generative term of the bound, $\vh^{(m)} \sim q(\vh|\vx)$ are $M$ independent samples drawn from the approximate posterior, and $\HH(q)$ is the entropy of the approximate posterior.

The most notable advances in variational inference were made in ``Helmholtz machines'' \cite{dayan1995helmholtz} that model the approximate posterior and conditional distributions by deep neural networks~\cite{mnih2014neural, bornschein2014reweighted}.  In this model, the difficulty is offset from inference to training the approximate posterior modeled by the ``recognition network''.

For example, suppose the conditional distribution is modeled by an FFN, such that the output makes up the parameters of a multivariate Gaussian distribution with mean, $\vmu_x$, and diagonal covariance, $\vsig_x$. Let us assume as well that the approximate posterior has a Logistic distribution with mean, $\vmu_h$, and scale, $\vs_h$:
\begin{align}
(\vmu_x(\vh; \TT), \vsig_x(\vh; \TT)) = g(\vh; \TT), \
(\vmu_h(\vx; \PS), \vs_h(\vx; \PS)) = f(\vx; \PS),
\end{align}
where $f$ and $g$ are multilayer FFNs with parameters $\TT$ and $\PS$, $\vx$ are visible variables corresponding to data, and $\vh$ are the latent variables (or sources). Finally, assume $p(\vh)$, the prior distribution of the latent variables, is a spherical multivariate Logistic distribution. The lower bound in Equation \ref{eq:approx_logp} becomes:
\begin{align}
\LL(\vx) &\approx \sum_{m=1}^M \bigg[ \log p(\vx | \vh^{(m)} ; \TT) \nonumber \\
&+ \sum_{i=1}^N \bigg( h_i^{(m)} - 2 \log \big(1 + \exp(h_i^{(m)}) \big) - \frac{h_i^{(m)} - \mu_h(\vx; \PS)_i}{s_h(\vx; \PS)_i} \nonumber \\
&+ \log s_h(\vx; \PS)_i + 2 \log\bigg(1 + \exp\bigg(\frac{h^{(m)}_i - \mu_h(\vx; \PS)_i}{s_h(\vx; \PS)_i}\bigg)\bigg) \bigg) \bigg]
\end{align}
The gradient of the first term above w.r.t the variational parameters, $\PS$, is not normally possible due to the stochastic variables $\vh^{(m)} \sim q(\vh | \vx)$. However, in the case of Logistic latent variables, the following re-parameterization makes learning possible via back-propagation:
\begin{align}
\vh = \vmu_h + \log\bigg(\frac{\vepsilon}{1 - \vepsilon}\bigg) \odot \vs_h \nonumber, \
\vepsilon \sim \mathcal{U}(0, 1).
\end{align}
Commonly known as a \emph{variational autoencoder} (VAE) \cite{mnih2014neural}, this type of re-parameterization is available for a number of continuous distributions, such as Gaussian, Poisson, and Gumbel, but is not available for Helmholtz machines with discrete latent variables, though other good methods exist~\cite{bornschein2014reweighted, hjelm2015iterative}. As the prior is factorized, the lower bound corresponds to learning a generative model with maximally-independent latent variables, a feature desirable in many research settings. This approach should, in principle, work for any directed graph with continuous latent variables, given the appropriate approximate posterior and prior.

%Undirected graphical models, such as restricted Boltzmann machines (RBMs) have been shown to be effective nonlinear models in medical imaging settings~\cite{hjelm2014restricted}. However, training deeper models rely heavily on pretraining, and in many cases results can be non-unique. In addition, sampling is difficult due to the intractable partition function, while sampling in directed graphical models can be done easily via ancestral sampling.

\vspace{-10pt}
\subsection{Visualizing Latent Variables}
\label{sec:viz}
%\vspace{-8pt}
Visualizing a latent variable, $h_i$, of a directed belief network involves calculating the marginal over all other latent variables: $p(\vx | h_i) = \sum_{\vh_{j \neq i}} p(\vx , \vh_{j \neq i} | h_i).$ This is computationally infeasible with most configurations. Alternatively, we can draw $M$ samples from the approximate posterior, $\vh_{j \neq i}^{(m)} \sim q(\vh_{j \neq i} | \vx)$ to approximate the marginal:
\begin{align}
p(\vx | h_i) \approx \sum_{m=1}^M \frac{p(\vx , \vh_{j \neq i} | h_i)}{q(\vh_{j \neq i} | \vx)}.
\end{align}
However, this approximation typically requires a large number of samples to be accurate (e.g. $O(100,000)$ with the MNIST dataset).  In addition, this only provides a single point in the marginal, which is a continuous function of $h_i$.  In reality, we are interested in how changes in $h_i$ effect generation of the image. Therefore, we use the following fast approximation to determine the ``projection" of the $i$th latent variable:
\begin{align}
\Delta p(\vx | h_i) / \Delta h_i \approx p(\vx | h_i = \mu_i + s_i, \vh_{j \neq i} = \vmu_{j \neq i}) - p(\vx | \vh = \vmu),
\end{align}
where $\mu_i$ and $s_i$ is reserved for parameters of the prior distribution that encode first and second order statistics respectively. For instance, for a Logistic distribution, $\mu_i$ would be the center of unit $i$ and $s_i$ would be the scale factor. This approximation does not capture the full generative effect of the latent variables, but it is sufficient for this demonstration.

\section{Experimental Setting}
For our medical imaging study, we used the MRI dataset from the combined schizophrenia studies in Plis et. al. \cite{plis2013deep}. Whole brain MRIs were obtained on a 1.5T Signa GE scanner using identical parameters and software, and the resulting dataset was segmented into grey matter regions with $60465$ voxels in each sample. For quality control, the correlation coefficient of each MRI volume was calculated and any volumes with mean coefficient of $2$ standard deviations below the mean across all volumes were categorized as noisy and removed. The resulting dataset had $163$ subjects and $156$ healthy controls.

For our generative model, we used a logistic prior, $p(\vh; \PP)$, with $64$ units and a 2-layer ``generation" feed-forward network (FFN) with a deterministic intermediate layer with $500$ softplus ($\log(1 + \exp(x))$) units to parameterize a Gaussian conditional distribution, $p(\vx | \vh; \TT)$. Our approximate posterior, $q(\vh | \vx; \PS)$, was a multivariate factorized logistic which was parameterized by a 2-layer ``recognition" FFN with $500$ hyperbolic tangent ($\tanh$) deterministic units. We learn $\PS$, $\TT$, and $\PP$ by maximizing the variational lower bound, and trained our model with a batch size of $10$ using the RMSprop algorithm~\cite{Hinton-Coursera2012} for $1000$ epochs. 
% Early experimental results were highly dependent on train-test-validation split, despite removing noisy samples; future experiments should use cross-validation, but to simplify our experiments we used the full set for training and used L2 weight decay with a coefficient of $0.0002$ on both the inference and the generation nets to avoid overfitting, as per results from early experiments. Early experiments also showed that VAE with Gaussian visible variables can have very unstable learning, however a learning rate of $0.0001$ stabilized it.

\section{Results}
\begin{figure}[]
\vspace{-25pt}
\begin{center}
\makebox[\textwidth] {\includegraphics[width=\textwidth]{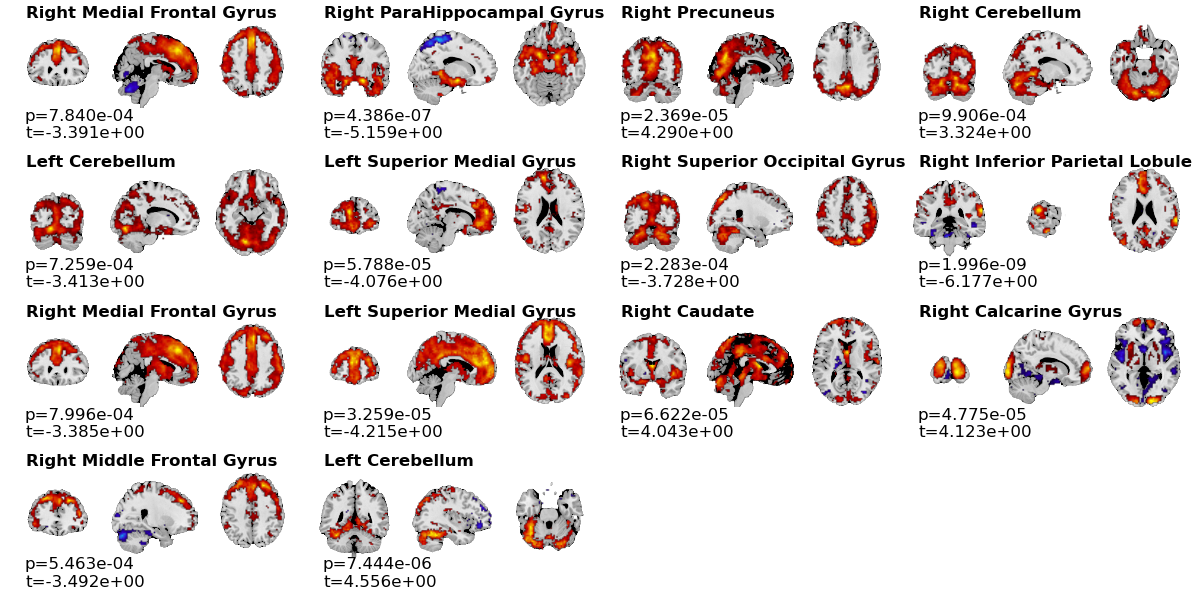}}
\end{center}
\vspace{-15pt}
\caption{
Projections of logistic latent variables for a variational autoencoder with an additional deterministic nonlinearity for the generative and recognition networks. Each projection was thresholded at $2$ standard deviations, and the latent variables showed here are those that showed high significance ($p < 0.001$) from a one-sampled $t$-test of the $\beta$ values from logistic regression to schizophrenia.}
\label{fig:features}
\vspace{-15pt}
\end{figure}
As the latent variable projections from Section \ref{sec:viz} were both positive and negative and the prior distribution is symmetric with respect to our choice of positive scale factors, we reversed the sign of our projections if the mean of voxels above $2$ standard deviations was negative. For each latent variable or ``component", we calculate the approximate posterior for each subject, $q(h_i | \vx_n)$, and then used logistic regression to schizophrenia using the approximate posterior means, $\vmu_h(\vx_n)$. Each component was tested for significance by using the resulting $\beta$ values from the logistic regression in a one-sample $t$-test. 

Visual inspection of the latent variables revealed a diverse set of features that were mostly identifiable as regions of interest, with very little noisy features. There was significantly more overlap between features than is typical with ICA with PCA preprocessing or RBM with MRI data \cite{hjelm2014restricted}, which may or may not be beneficial depending on the research setting. Latent variables that showed high significance to schizophrenia ($p < 0.001$) are shown in Figure \ref{fig:features} with the complete set in the Supplementary Material.
\begin{figure}
\centering
\begin{minipage}{.48\textwidth}
  \centering
  \includegraphics[width= .9\textwidth]{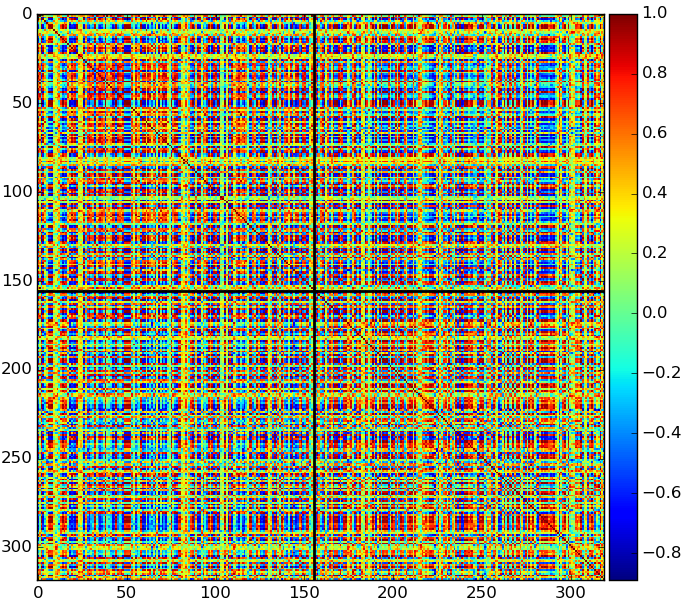}
  \captionof{figure}{The correlation matrix of the logistic centers of the approximate posterior for each subject, $\vmu_q(\vx_n)$ for components with high significance ($p < 0.001$) to schizophrenia. Columns and rows have been ordered to show healthy controls first, followed by patients, with higher inter-group correlation.}
  \label{fig:s_corr}
\end{minipage}%
\hspace{1em}
\begin{minipage}{.48\textwidth}
  \centering
  \vspace{-10pt}
  \includegraphics[width= .9\textwidth]{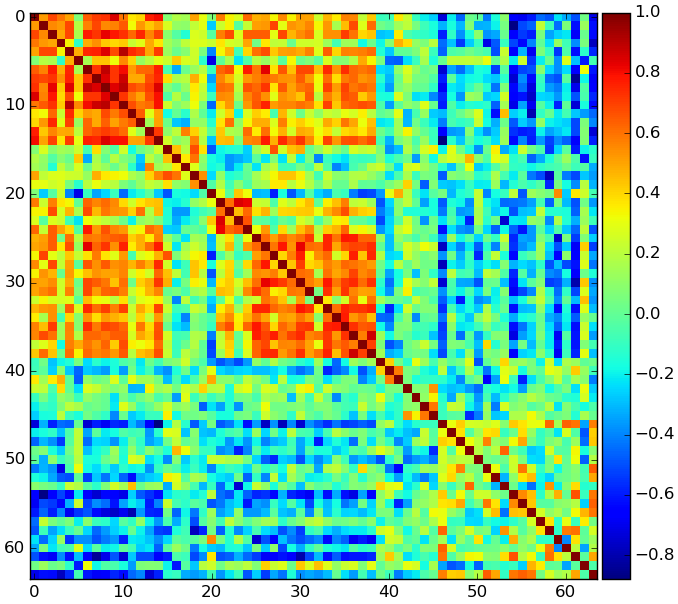}
  \captionof{figure}{Correlation matrix between all components across subjects. Rows and columns were ordered according to grouping determined by a community multi-level analysis \cite{blondel2008fast} and shows inter- and intra-group structure between components.}
  \label{fig:comp_corr}
\end{minipage}
\vspace{-20pt}
\end{figure}
The means of the approximate posterior, $\vmu_h(\vx_n)$, were used as input to a classification task, using simple logistic regression and $100$-fold class-balanced cross validation. The resulting classification rate, $0.67$, is  significantly above chance. The conclusion is that, despite lacking information about the labels in the MLE objective and much lower dimensionality, the latent variables have a similar amount of information necessary to perform diagnosis. This is also apparent in the correlation matrix in Figure \ref{fig:s_corr} using the components that showed high significance to schizophrenia. Finally, the components were grouped by calculating the correlation of approximate posterior centers across subjects. Figure \ref{fig:comp_corr} shows several groupings, as well as some inter-group relationships.

\vspace{-10pt}
\section{Conclusions}
\vspace{-10pt}
We have demonstrated variational autoencoders as a means of training nonlinear directed graphical models for extracting maximally indepdent features from MRI data. Our results show both relevant structure and preservation of information relevant to schizophrenia diagnosis. This work opens the door for further studies using Helmholtz machines for medical imaging research, including multimodal and multilayer analysis.

\vspace{-10pt}
{\small
\bibliography{miccai}
}
\bibliographystyle{splncsnat}

\end{document}